\documentclass{article}

\usepackage{microtype}
\usepackage{graphicx}
\usepackage{subcaption}
\usepackage{booktabs}

\usepackage{hyperref}

\usepackage[preprint]{icml2026}

\usepackage{amsmath}
\usepackage{amssymb}
\usepackage{mathtools}
\usepackage{amsthm}
\usepackage{enumitem}
\usepackage{float}

\usepackage[capitalize,noabbrev]{cleveref}

\theoremstyle{plain}

\theoremstyle{definition}

\theoremstyle{remark}

\usepackage[textsize=tiny]{todonotes}

\icmltitlerunning{Layer-Resolved Optimal Transport for Hallucination Detection}

\begin{document}

\twocolumn[
  \icmltitle{Layer-Resolved Optimal Transport for Hallucination Detection \\
             in NMT and Abstractive Summarization}

  \begin{icmlauthorlist}
    \icmlauthor{Mariia Onyshchuk}{ucu}
    \icmlauthor{Maksym-Vasyl Tarnavskyi}{ucu}
    \icmlauthor{Marta Sumyk}{ucu}
  \end{icmlauthorlist}

  \icmlaffiliation{ucu}{Faculty of Applied Sciences, Ukrainian Catholic University, Lviv, Ukraine}

  \icmlcorrespondingauthor{Mariia Onyshchuk}{onyshchuk.pn@ucu.edu.ua}
  \icmlcorrespondingauthor{Maksym-Vasyl Tarnavskyi}{tarnavskyi.pn@ucu.edu.ua}
  \icmlcorrespondingauthor{Marta Sumyk}{sumyk.pn@ucu.edu.ua}

  \icmlkeywords{Optimal Transport, Hallucination Detection, Neural Machine Translation, Abstractive Summarization, Cross-Attention Analysis}

  \vskip 0.3in
]

\printAffiliationsAndNotice{Accepted to the \href{https://mechinterpworkshop.com/}{Mechanistic Interpretability Workshop}.}

% --  --  --  --  --  --  --  --  --  --  --  --  --  --  --  --  --  --  --  --  --  --  --  --  --  --  --  --  -- -
%	ABSTRACT
% --  --  --  --  --  --  --  --  --  --  --  --  --  --  --  --  --  --  --  --  --  --  --  --  --  --  --  --  -- -
\begin{abstract}
Optimal transport (OT) has been shown to detect hallucinations in neural machine translation (NMT) by measuring the geometric distance between cross-attention distributions and a reference distribution, without any supervision~\citet{guerreiro2022}. We extend this analysis to all six decoder layers of the Fairseq DE-EN model ($N=3{,}414$), showing that Wass-to-Unif and Wass-to-Data are complementary detectors specialised across hallucination types, that detection is concentrated in layers L1--L4 with L5 anti-predictive for subtler types, and that hallucinated translations lack the exploratory attention phase present in correct translations from the first decoding step. We further evaluate whether the geometric signal transfers to abstractive summarization faithfulness detection: our unsupervised OT detector on AggreFact~\citet{tang2023aggrefact} ($N=1{,}116$) achieves $57.2\%$/$57.6\%$ balanced accuracy on CNN/XSum -- above chance but substantially below supervised MiniCheck-Flan-T5-L~\citet{tang2023minicheck} ($69.9\%$/$74.3\%$). This gap is principled: unlike NMT hallucinations, unfaithful summaries can attend correctly to source tokens while misrepresenting their content, a failure mode invisible to concentration-based OT metrics by construction. Structural experiments on T5-base~\citet{raffel2020} confirm consistent decoder organisation across depth, with Layer~3 showing peak concentration and Layer~12 being most critical for generation quality. Together, the results establish OT on cross-attention as a reliable detector when the failure mode is source disengagement, a principled interpretability tool regardless of task, and fundamentally limited when faithfulness failures occur downstream of attention.
\end{abstract}

% --  --  --  --  --  --  --  --  --  --  --  --  --  --  --  --  --  --  --  --  --  --  --  --  --  --  --  --  -- -
%	1. INTRODUCTION
% --  --  --  --  --  --  --  --  --  --  --  --  --  --  --  --  --  --  --  --  --  --  --  --  --  --  --  --  -- -
\section{Introduction}

Transformer models~\citet{vaswani2017attention} have achieved strong performance
in abstractive summarization, yet their internal attention mechanisms
remain poorly understood. A practical concern is faithfulness: models
can generate summaries that are fluent but factually inconsistent with
the source, a failure mode closely related to hallucination in neural
machine translation (NMT). Recent work~\citet{maynez2020faithfulness,kryscinski2020evaluating}
has shown that faithfulness failures are common even in state-of-the-art
summarization systems.

\citet{guerreiro2022} demonstrated that hallucinations in NMT produce
cross-attention distributions that are geometrically detached from the
source, and that this detachment is measurable via Wasserstein-1 ($W_1$)
distance. Their fully unsupervised detector outperformed all prior
model-based approaches and was competitive with external models trained
on millions of samples for quality estimation and cross-lingual sentence
similarity. However, their analysis operates on a single aggregate
attention distribution from the final decoder layer, leaving open how
the hallucination signal distributes across individual layers, how
different hallucination types relate to different detectors, and
whether the geometric intuition transfers beyond NMT.

We address both of these open questions. First, we extend the original
NMT analysis to all six decoder layers of the Fairseq DE-EN model,
introducing routing consistency as an additional detector and
characterising the layer-resolved geometry of each hallucination type.
Second, we ask whether the geometric signal transfers to abstractive
summarization faithfulness detection, using the T5 architecture~\citet{raffel2020}
as a testbed and evaluating on the AggreFact benchmark~\citet{tang2023aggrefact}.

Our contributions are:

\begin{enumerate}

\item A layer-resolved analysis of the Fairseq DE-EN hallucination
corpus of~\citet{guerreiro2022}, extending their aggregate last-layer
result to all six decoder layers and introducing routing consistency
as an additional unsupervised detector. We show that Wass-to-Unif
and Wass-to-Data are complementary detectors specialised across
hallucination types, that detection performance is concentrated in
layers L1--L4, and that hallucinated translations lack the exploratory
attention phase present in correct translations from the first decoding
step.

\item The first application of OT-based hallucination detection to
abstractive summarization, evaluated on the AggreFact
benchmark~\citet{tang2023aggrefact} against supervised baselines
including MiniCheck~\citet{tang2023minicheck}.

\item A theoretical account of why the NMT-to-summarization transfer
is partial, grounding the empirical gap in the distinction between
retrieval failure and content misuse, and calibrating it against the
gradient of detectability observed across NMT hallucination types.

\item A structural analysis of T5-base cross-attention geometry across
all 12 decoder layers via OT metrics, revealing consistent
architectural organisation confirmed by leave-one-out ablation and
convergent with the layer structure identified in the Fairseq model.

\end{enumerate}

\section{Background}

\subsection{OT-Based Hallucination Detection in NMT}

Given two discrete probability distributions $\mu$ and $\nu$ over positions
$\{1,\ldots,S\}$, the Wasserstein-1 distance is defined as:
\begin{equation}
  W_1(\mu,\nu)
  \;=\;
  \inf_{\gamma \,\in\, \Gamma(\mu,\nu)}
  \int_{\mathbb{R}\times\mathbb{R}} |x - y|\; d\gamma(x,y),
  \label{eq:w1_def}
\end{equation}
where $\Gamma(\mu,\nu)$ is the set of all joint distributions (transport plans) with
marginals $\mu$ and $\nu$, and $|x-y|$ is the ground metric on token positions.
Intuitively, $W_1$ measures the minimum ``work'' needed to rearrange one distribution
into the other, making it sensitive to the spatial structure of attention mass in a way
that entropy-based measures are not.
For discrete distributions on a 1D grid, $W_1$ reduces to the area between cumulative
distribution functions, enabling efficient exact computation \citet{peyre2019computational}.

\citet{guerreiro2022} proposed treating each cross-attention distribution
as a point in the space of probability measures over source positions, and measuring
its concentration via W$_1$ distance to the uniform distribution
$\mathbf{u} = (1/S, \ldots, 1/S)^\top$:
\begin{equation}
  c^{(\ell,t)} = W_1\!\left(\pi^{(\ell,t)},\, \mathbf{u}\right),
  \label{eq:wtu}
\end{equation}
where $\pi^{(\ell,t)}$ is the head-averaged cross-attention distribution at decoder layer
$\ell$ and generation step $t$.
Low concentration -- attention mass spread uniformly across source positions -- flags
potential hallucination.
Their per-example score aggregates this signal as the layer-median mean of
$c^{(\ell,t)}$, and their Wass-to-Unif (WTU) and Wass-to-Data (WTD) detectors are
complementary: WTU captures absolute concentration while WTD measures distributional
similarity to a reference set of confirmed-correct translations.

\subsection{Faithfulness in Summarisation}

Faithfulness failures in abstractive summarisation differ fundamentally from
NMT hallucinations~\citet{maynez2020faithfulness}.
\citet{maynez2020faithfulness} distinguish \emph{intrinsic} hallucinations, where
generated content contradicts the source, from \emph{extrinsic} ones, where content
cannot be verified from the source.
NMT hallucinations are predominantly intrinsic and severe -- the decoder ignores
source content almost entirely.
Abstractive summarisation failures are often extrinsic: the model attends correctly
to source tokens but infers or distorts beyond what is licensed by the evidence.
This distinction is central to understanding why OT transfer is partial: the signal
that works in NMT (source disengagement) is simply not the dominant failure mode
in abstractive summarisation.

\subsection{Attention as an Interpretability Signal}

Cross-attention distributions in encoder--decoder Transformers have been used as
a proxy for source--target alignment~\citet{vaswani2017attention}.
More recently, mechanistic interpretability work has identified functional
specialisation across decoder layers.
OT provides a principled, label-free way to characterise this structure via the
geometry of attention distributions, without requiring probing classifiers or
task-specific supervision.

%% ============================================================
\section{Methodology}

\subsection{Attention Extraction}

For each source--output pair we extract the full cross-attention tensors from the
decoder.
For layer $\ell$ and generation step $t$, the raw tensor has shape $(H, T_{\text{tgt}}, S)$,
where $H$ is the number of heads, $T_{\text{tgt}}$ the output length, and $S$ the source
length.
We average over heads:
\begin{equation}
  \pi^{(\ell,t)} = \frac{1}{H}\sum_{h=1}^{H} \alpha^{(h,\ell,t)} \in \Delta^{S-1},
\end{equation}
following \citet{guerreiro2022}.

\subsection{OT Metrics}

\paragraph{Wass-to-Unif (WTU).}
The per-layer concentration score is the mean W$_1$ distance to the uniform
distribution over decoding steps:
\begin{equation}
  s^{(\ell)}_{\text{WTU}} = \frac{1}{T}\sum_{t=1}^{T} W_1\!\left(\pi^{(\ell,t)}, \mathbf{u}\right).
\end{equation}
The aggregate per-example score averages over layers: low scores flag potential
hallucination.

\paragraph{Step-to-step OT.}
To measure how dynamically the decoder repositions its source attention
during generation, we compute the $W_1$ distance between attention
distributions at consecutive steps within each layer:
\begin{equation}
  [\mathbf{S}]_{\ell,t} = W_1\!\left(\pi^{(\ell,t)},\, \pi^{(\ell,t+1)}\right),
  \quad t = 1,\ldots,T-1.
\end{equation}
The per-layer mean step-OT summarises the average attention shift. A
model that scans the source dynamically produces high step-OT; one that
locks onto fixed positions produces low step-OT.

\paragraph{Layer-pairwise OT.}
To characterise routing similarity across decoder depth, we compute $W_1$
between step-averaged attention distributions at each pair of layers:
\begin{equation}
  [\mathbf{D}]_{\ell,\ell'} = W_1\!\left(\bar{\pi}^{(\ell)},\,
      \bar{\pi}^{(\ell')}\right), \quad
  \bar{\pi}^{(\ell)} = \frac{1}{T}\sum_{t=1}^{T} \pi^{(\ell,t)},
\end{equation}
producing a symmetric $L \times L$ distance matrix per example. Low
$[\mathbf{D}]_{\ell,\ell'}$ indicates that two layers attend to similar
source positions on average; high values indicate functionally distinct
routing.

\paragraph{Wass-to-Data (WTD).}
For each test sentence we retrieve $k{=}4$ nearest neighbours from a reference set
of confirmed-correct sentences (filtered by length proximity $\delta{=}0.1$) and
compute the mean W$_1$ distance to their step-averaged attention distributions.

\paragraph{Routing Consistency (RC).}
Let $\hat{\jmath}(\ell, t) = \arg\max_j \alpha^{(\ell,t)}_j$ be the argmax source
position at step $t$.
The routing entropy at layer $\ell$ is
$H(\ell) = -\sum_j \hat{p}^{(\ell)}_j \log \hat{p}^{(\ell)}_j$,
where $\hat{p}^{(\ell)}_j = T^{-1}\sum_t \mathbf{1}[\hat{\jmath}(\ell,t)=j]$.
The per-example RC score is $-L^{-1}\sum_\ell H(\ell)$, negated so that higher
values correspond to more consistent, focused routing.
RC captures a complementary aspect of attention geometry: not how concentrated
the distribution is, but how consistently the decoder returns to the same source
position across steps.

All W$_1$ distances are computed exactly via the CDF formula~\citet{peyre2019computational},
avoiding Sinkhorn regularisation error.

\subsection{Datasets}

\paragraph{NMT (Fairseq DE-EN).}
We use the annotated corpus of \citet{guerreiro2022}: 3,414 DE$\to$EN
translations with binary labels for five hallucination categories.
Following their taxonomy, we define the hallucination group as sentences positive
for any of full-unsupport (129), strong-unsupport (164), or repetitions (87) -- yielding
324 hallucinated sentences -- and the confirmed-correct group as 2,882 sentences
where all five label columns are zero.

\paragraph{Summarisation (AggreFact).}
We evaluate on the AggreFact benchmark~\citet{tang2023aggrefact}, using test
splits for CNN/DailyMail ($N{=}558$) and XSum ($N{=}558$).
The primary supervised baseline is MiniCheck-Flan-T5-L~\footnote{\url{https://huggingface.co/lytang/MiniCheck-Flan-T5-Large}}
(0.8B parameters), which achieves 69.9\%/74.3\% balanced accuracy on CNN/XSum.
Balanced accuracy (BAcc), the arithmetic mean of sensitivity and specificity, is used
as the primary metric to account for class imbalance (CNN: 89.8\% faithful).
Structural experiments use T5-base~\footnote{\url{https://huggingface.co/google-t5/t5-base}} ($N{=}100$ CNN/DailyMail
examples for concentration profiling; $N{=}50$ for quality-group comparisons).

%% ============================================================
\section{Experiments I: Revisiting Translation}
\label{sec:nmt}

\subsection{Layer Structure and Concentration Profile}

\begin{figure}[ht]
    \centering
    \includegraphics[width=0.55\linewidth]{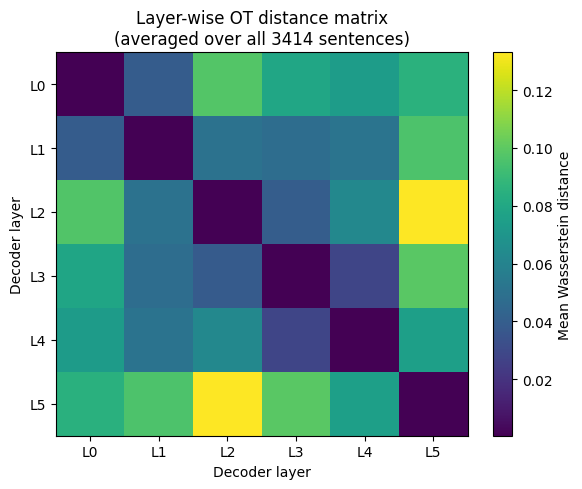}
    \caption{Mean pairwise $W_1$ distance matrix $\mathbf{D}$ averaged
    over all 3{,}414 sentences (Fairseq DE-EN). Three functional regimes
    are visible: L0 (transitional), L1--L4 (tight cluster, L1 being bridging to L0),
    and L5 (isolated; maximum distance $\approx 0.13$ from L2).}
    \label{fig:layer-ot-heatmap}
\end{figure}

Figure~\ref{fig:layer-ot-heatmap} shows the mean pairwise $W_1$ distance
matrix $\mathbf{D}$ averaged over all 3{,}414 sentences. The matrix reveals
three distinct functional regimes. L0 is moderately distant from all other
layers, with no strong affinity to any particular group. L1 acts as a
transitional layer: its row is dark toward both L0 and the central block,
indicating it shares routing behaviour with both neighbours rather than
belonging cleanly to either. Layers L2--L4 form a tight cluster, with
pairwise distances among themselves substantially lower than to any other
layer, suggesting shared or functionally redundant source-routing behaviour.
Finally, L5 is structurally isolated: the maximum pairwise distance in the
entire matrix occurs between L2 and L5 (mean $W_1 \approx 0.13$, yellow
cell), and every entry in L5's row is markedly brighter than the interior
of the L2--L4 block. This is consistent with L5's role as the final decoder
layer, which must commit to specific source tokens immediately before
generation and therefore implements qualitatively different routing from the
preceding layers.

\begin{figure}[ht]
    \centering
    \includegraphics[width=0.55\linewidth]{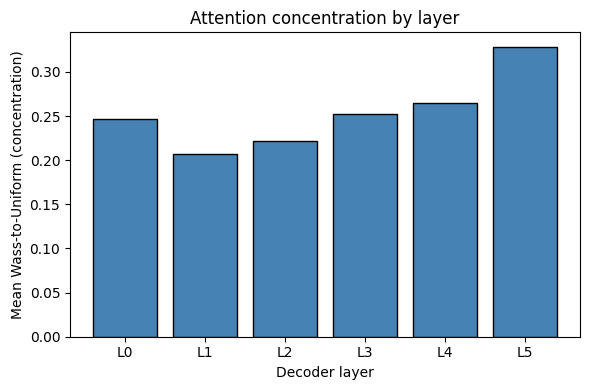}
    \caption{Mean $W_1$ concentration $s_{\mathrm{WTU}}^{(\ell)}$ per
    decoder layer (Fairseq DE-EN, $N=3{,}414$). The profile increases
    broadly from L1 ($0.21$) to L5 ($0.33$), with the sharpest jump at
    the final layer. L0 sits slightly above L1, producing a minor
    non-monotonicity at the decoder entry.}
    \label{fig:concentration-bar}
\end{figure}

Figure~\ref{fig:concentration-bar} shows the mean concentration
$s_{\mathrm{WTU}}^{(\ell)}$ per layer. The profile is broadly monotone
increasing from L1 ($0.21$) through L5 ($0.33$), with L0 sitting slightly
above L1 at $0.25$ -- a minor non-monotonicity at the decoder entry. The
sharpest single jump occurs between L4 ($0.26$) and L5 ($0.33$), consistent
with L5's structural isolation: it routes differently from all other layers
and does so by attending more selectively than any of them.

\medskip
\begin{minipage}{\linewidth}
\noindent\textbf{Finding 1.} \textit{The six Fairseq decoder layers
organise into three functional regimes: L0 is transitional, L1 bridges
early and middle layers, L2--L4 form a tight cluster with shared routing
behaviour, and L5 is structurally isolated as the most concentrated and
routing-divergent layer in the network.}
\end{minipage}
\subsection{Replication and Hallucination Separation}

\begin{table*}[ht]
    \centering
    \caption{Mann-Whitney $U$ test: hallucinated vs.\ confirmed-correct
    translations (Fairseq DE-EN, $N_{\text{hall}}{=}324$,
    $N_{\text{corr}}{=}2{,}882$). All differences significant at
    $p{<}0.001$. The correct group excludes 208 sentences with
    non-hallucination error labels.}
    \label{tab:mannwhitney}
    \begin{tabular}{lrrr}
        \toprule
        Metric & Hall.\ mean & Correct mean & $p$-value \\
        \midrule
        Mean step OT              & 0.082 & 0.124 & $<0.001$ \\
        Std.\ concentration       & 0.090 & 0.104 & $<0.001$ \\
        Mean concentration        & 0.296 & 0.249 & $<0.001$ \\
        First-layer concentration & 0.257 & 0.246 & $<0.001$ \\
        Mean layer OT             & 0.055 & 0.059 & $<0.001$ \\
        Final-layer concentration & 0.338 & 0.328 & $<0.001$ \\
        \bottomrule
    \end{tabular}
\end{table*}

Table~\ref{tab:mannwhitney} confirms the geometric signal identified by
\citet{guerreiro2022} with exact $W_1$ distances computed at each
individual decoder layer. Hallucinated translations exhibit significantly
higher mean concentration ($0.296$ vs.\ $0.249$, $p{<}0.001$,
Mann-Whitney~U): attention mass is more tightly focused on a sparse set
of source tokens throughout generation. They also show significantly lower
step-to-step OT ($0.082$ vs.\ $0.124$, $p{<}0.001$): a hallucinating
decoder locks onto fixed source positions and stays there, while a
correctly translating model dynamically scans as it generates. This
\emph{static attention} signature -- not reported in the original paper,
which did not compute step-resolved OT -- provides a complementary
characterisation of the failure mode beyond concentration alone.

The remaining metrics reinforce this picture. Lower standard deviation
of concentration ($0.090$ vs.\ $0.104$) confirms that hallucinated
attention is not only more peaked on average but also less variable
across steps. Lower mean layer OT ($0.055$ vs.\ $0.059$) indicates that
hallucinated sentences exhibit less divergence across decoder layers,
consistent with a model that has committed to an output independently
of source content and does so uniformly across the stack.

\begin{figure}[ht]
    \centering
    \includegraphics[width=\linewidth]{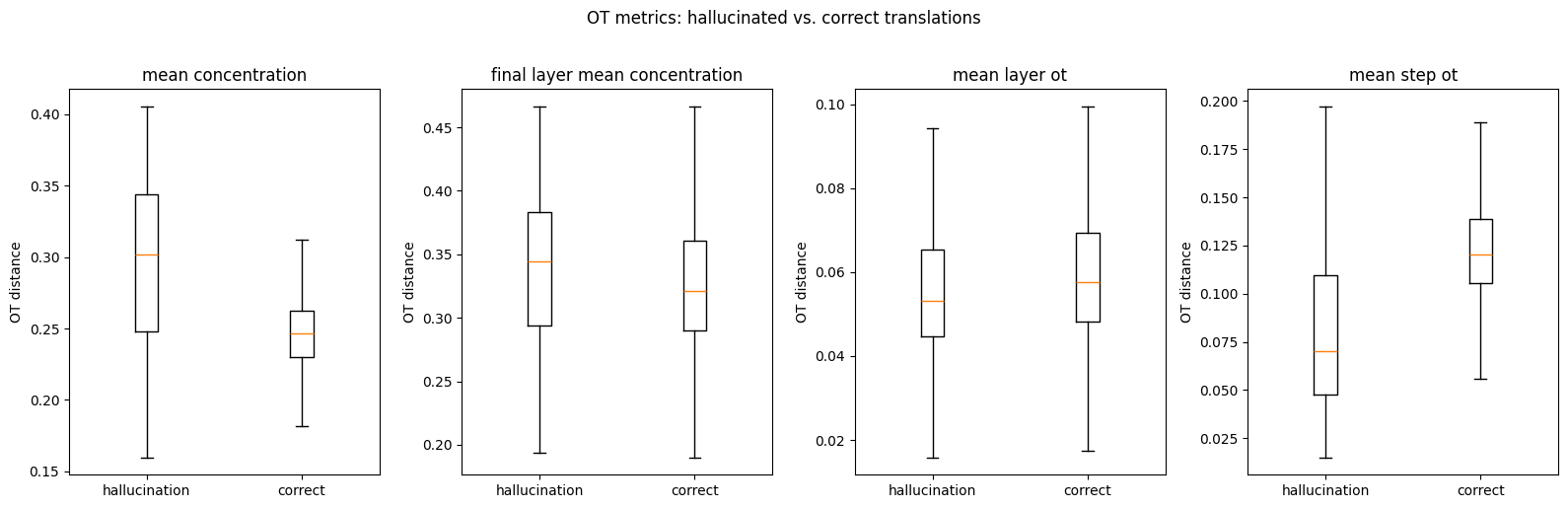}
    \caption{OT metrics for hallucinated vs.\ confirmed-correct
    translations (Fairseq DE-EN, $N{=}324$ hallucinated,
    $N{=}2{,}882$ correct; outliers suppressed). Left to right: mean
    concentration, final-layer concentration, mean layer OT, mean step
    OT. All differences significant at $p{<}0.001$ (Mann-Whitney $U$).}
    \label{fig:boxplots-4panel}
\end{figure}

Figure~\ref{fig:boxplots-4panel} shows the four most interpretable metrics
as boxplots. The mean concentration panel shows the strongest visual
separation: the hallucination box sits almost entirely above the correct
box with minimal interquartile overlap. The step-OT panel shows the inverse
pattern with comparable clarity. Final-layer concentration and mean layer
OT show significant but weaker separation, consistent with their smaller
absolute differences in Table~\ref{tab:mannwhitney}.

\medskip
\noindent\textbf{Finding 2.} \textit{Hallucinated translations are
distinguished from correct translations on every OT metric
(all $p{<}0.001$). The two most discriminative signals are higher mean
concentration ($+0.047$) and lower step-to-step OT ($-0.042$), jointly
characterising hallucination as static, focused attention that locks onto
irrelevant source positions from the first decoding step.}

\subsection{Layer-Resolved Detection Performance}

\begin{figure}[ht]
    \centering
    \includegraphics[width=0.85\linewidth]{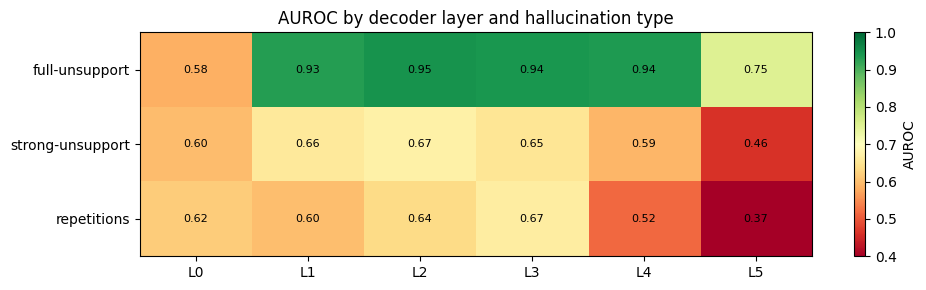}
    \caption{AUROC of the WTU detector by decoder layer and hallucination
    type (Fairseq DE-EN). The RdYlGn colormap spans $[0.4, 1.0]$; values
    below $0.5$ (red) indicate anti-predictive layers. Full-unsupport is
    strongly detectable at L1--L4; strong-unsupport and repetitions are
    weakly detectable throughout; L5 is anti-predictive for the two
    harder types.}
    \label{fig:auroc-heatmap}
\end{figure}

Figure~\ref{fig:auroc-heatmap} reports WTU AUROC by decoder layer and
hallucination type. Results are strongly type-dependent. For full-unsupport
hallucinations, WTU is a reliable detector: AUROC peaks at L2 ($0.946$)
and remains above $0.93$ for L1--L4, before dropping to $0.750$ at L5.
L0 is near-random ($0.584$), confirming that the earliest layer carries
almost no concentration-based hallucination signal. For strong-unsupport
and repetitions, performance is substantially lower throughout (maximum
per-layer AUROC $0.672$ and $0.667$ respectively).

A notable inversion appears at L5: WTU AUROC falls below chance for
strong-unsupport ($0.461$) and repetitions ($0.374$), meaning final-layer
concentration is actively \emph{anti-predictive} for these types. This
reflects the layer's structural role: L5 is always the most concentrated
layer regardless of translation quality (mean WTU $0.33$ corpus-wide),
so its concentration provides no discriminative signal for hallucination
types that retain some source routing.

\subsection{Routing Consistency as a New Detector}

RC outperforms WTU for full-unsupport both in aggregate ($0.957$
vs.\ $0.937$) and at its best layer, L2 ($0.955$ vs.\ $0.946$), making
it the single strongest individual detector in our analysis
(Table~\ref{tab:rc_auroc}). The explanation is mechanistic: full-unsupport
hallucinations have RC near zero at every layer -- the decoder routes to
the same source position at every step -- while confirmed-correct
translations dip to $-1.77$ at L2, reflecting highly diverse source
scanning. It is not merely that attention is concentrated on a wrong token;
it is that the \emph{same} token receives maximum attention at every step
regardless of what is being generated. For repetitions, both WTU and RC
are near chance ($0.568$ and $0.508$), confirming that oscillatory
hallucinations have a qualitatively different attention signature that
neither metric captures reliably.

\begin{table*}[ht]
\centering
\caption{Routing Consistency (RC) detector AUROC by decoder layer and
hallucination type (Fairseq DE-EN). Higher is better.}
\label{tab:rc_auroc}
\begin{tabular}{lrrrrrrr}
\toprule
Type & Agg. & L0 & L1 & L2 & L3 & L4 & L5 \\
\midrule
Full-unsupport   & 0.957 & 0.789 & 0.785 & \textbf{0.955} & 0.945 & 0.928 & 0.881 \\
Strong-unsupport & 0.672 & 0.653 & 0.620 & \textbf{0.694} & 0.689 & 0.626 & 0.572 \\
Repetitions      & 0.508 & 0.536 & 0.485 & 0.540 & 0.541 & 0.468 & 0.414 \\
\bottomrule
\end{tabular}
\end{table*}

\begin{figure}[ht]
    \centering
    \includegraphics[width=\linewidth]{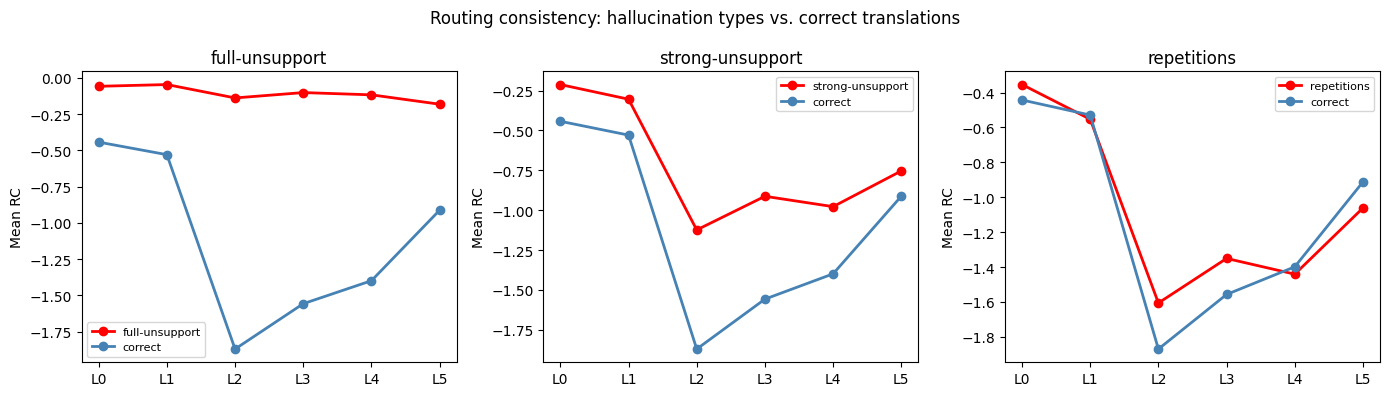}
    \caption{Mean routing consistency $\mathrm{RC}(\ell)$ per decoder
    layer, split by hallucination type and confirmed-correct translations
    (Fairseq DE-EN). Full-unsupport hallucinations (red, left panel) have
    RC near zero at every layer -- they route to a single source position
    throughout decoding -- while correct translations (blue) dip sharply
    at L2 ($-1.77$), reflecting diverse source routing. Strong-unsupport
    shows partial separation; repetitions are indistinguishable.}
    \label{fig:rc-by-type}
\end{figure}

\begin{table}[ht]
    \centering
    \caption{Aggregate WTD vs.\ WTU AUROC by hallucination type
    (Fairseq DE-EN). Bold indicates the stronger detector per type.}
    \label{tab:wtd-wtu}
    \begin{tabular}{lrr}
        \toprule
        Type & WTD & WTU \\
        \midrule
        Full-unsupport   & 0.801 & \textbf{0.937} \\
        Strong-unsupport & \textbf{0.770} & 0.629 \\
        Repetitions      & \textbf{0.790} & 0.568 \\
        \bottomrule
    \end{tabular}
\end{table}

\begin{figure}[ht]
    \centering
    \includegraphics[width=\linewidth]{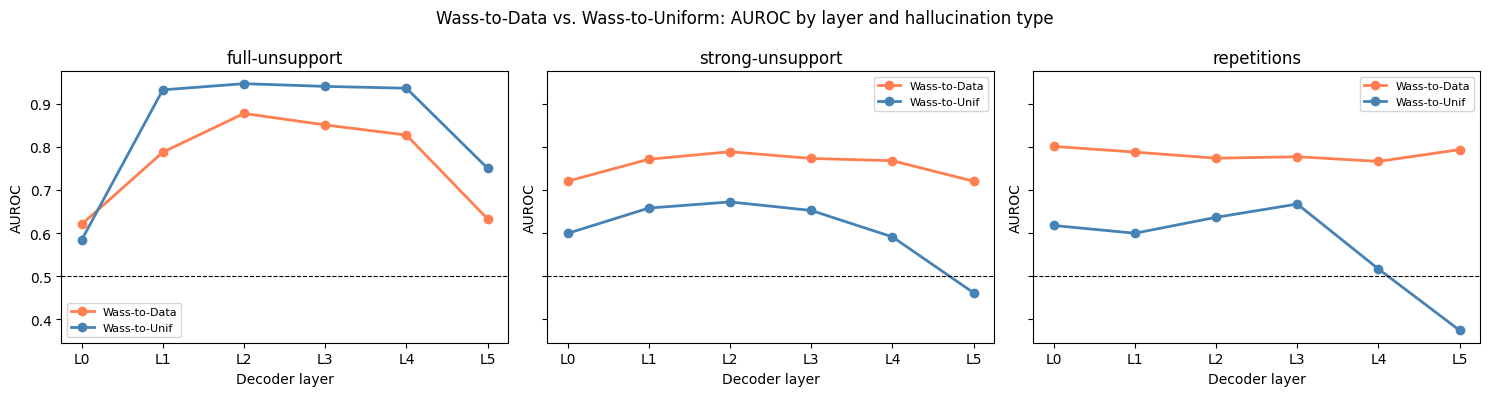}
    \caption{Per-layer AUROC for WTD (orange) and WTU (blue) by
    hallucination type (Fairseq DE-EN). WTU dominates for full-unsupport
    at all layers; WTD dominates for strong-unsupport and repetitions and
    is robust to the L5 collapse that afflicts WTU. The dashed line marks
    the random baseline ($0.5$).}
    \label{fig:wtd-wtu}
\end{figure}

WTD and WTU are complementary across hallucination types: WTU dominates
for full-unsupport ($0.937$ vs.\ WTD $0.801$) where absolute concentration
is the discriminative signal, while WTD dominates for strong-unsupport
($0.770$ vs.\ $0.629$) and repetitions ($0.790$ vs.\ $0.568$), where shape
similarity to reference distributions carries more information than
concentration alone. Figure~\ref{fig:wtd-wtu} shows the per-layer picture.
For strong-unsupport, WTD sits consistently above WTU across L0--L4
($0.72$--$0.79$ vs.\ $0.60$--$0.67$); at L5, WTU collapses below chance
($0.46$) while WTD holds steady at ${\sim}0.72$. The repetitions panel
shows the most extreme divergence: WTD is flat across all six layers
($0.79$--$0.81$), while WTU collapses catastrophically at L4--L5, reaching
$0.37$ at L5 -- actively misleading. This complementarity provides empirical
grounding for the Wass-Combo design of \citet{guerreiro2022}: the
two detectors are specialised at the level of individual layers and
hallucination types in consistent and interpretable ways.

\medskip
\noindent\textbf{Finding 3.} \textit{WTU and RC are strong detectors for
full-unsupport hallucinations (AUROC $0.937$ and $0.957$ respectively),
peaking at L2. WTD is the stronger detector for strong-unsupport and
repetitions (AUROC $0.770$ and $0.790$), robust across all layers including
L5 where WTU becomes anti-predictive. The three detectors are
complementary: WTU and RC capture absolute concentration and routing
diversity; WTD captures distributional shape relative to correct
translations.}

\subsection{Generation Dynamics}

Normal decoding follows a consistent two-phase trajectory across all six
decoder layers. In the \emph{exploration phase} (roughly the first $25\%$
of the sequence), attention shifts rapidly between source positions and
concentration is low -- the model is actively scanning the source. In the
\emph{commitment phase} (the remaining $75\%$), attention shifts decay and
concentration rises as the model progressively locks onto specific source
tokens.

Hallucinated translations lack this structure entirely. Layer trajectories
are compressed together from the very first decoding step, with no
pronounced early dip and little fan-out between layers: the model begins
in a statically concentrated state, skipping the exploratory phase before
generation properly starts. This confirms that the attention pathology
underlying hallucination is present from step~1, with a practical
implication: early-step OT scores could in principle support online
detection before the full output is generated.

\begin{figure}[ht]
    \centering
    \begin{subfigure}[b]{0.49\linewidth}
        \includegraphics[width=\linewidth]{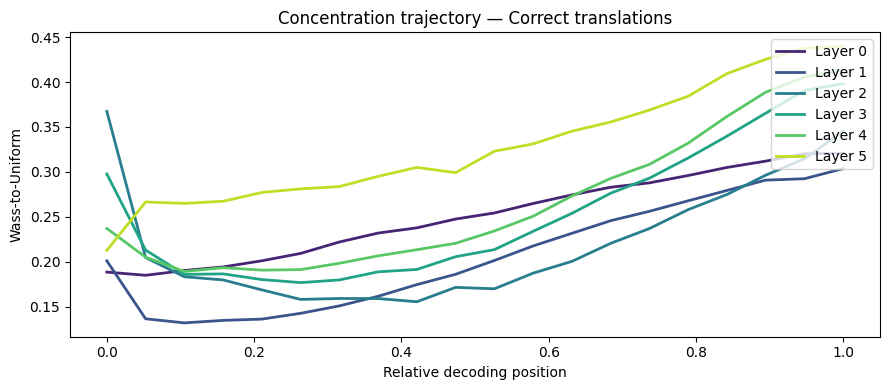}
        \caption{Concentration trajectory.}
    \end{subfigure}
    \hfill
    \begin{subfigure}[b]{0.49\linewidth}
        \includegraphics[width=\linewidth]{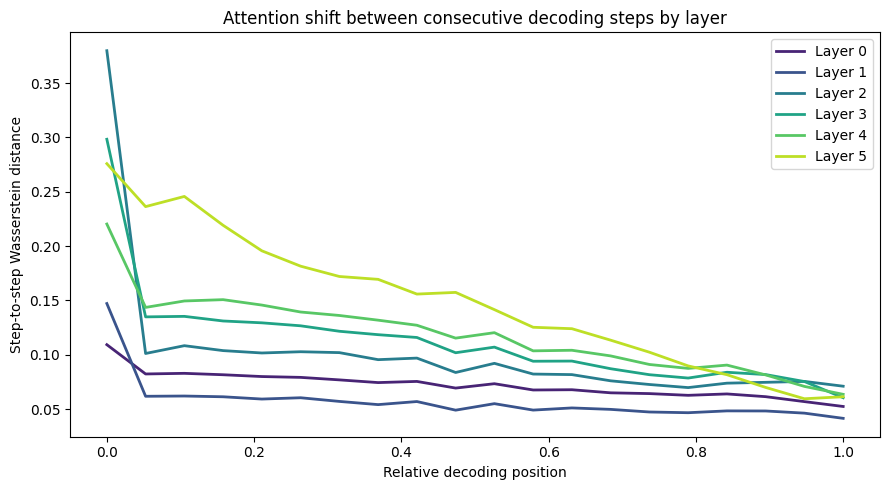}
        \caption{Step-to-step OT trajectory.}
    \end{subfigure}
    \caption{Mean concentration (left) and step-to-step $W_1$ distance
    (right) across relative decoding position, averaged over all 3{,}414
    sentences (Fairseq DE-EN). Both reveal a two-phase dynamic: early
    exploration (high shift, low concentration) followed by commitment
    (low shift, rising concentration).}
    \label{fig:step-trajectories}
\end{figure}

\begin{figure}[ht]
    \centering
    \begin{subfigure}[b]{0.49\linewidth}
        \includegraphics[width=\linewidth]{concentration_trajectory_correct.png}
        \caption{Confirmed-correct translations.}
    \end{subfigure}
    \hfill
    \begin{subfigure}[b]{0.49\linewidth}
        \includegraphics[width=\linewidth]{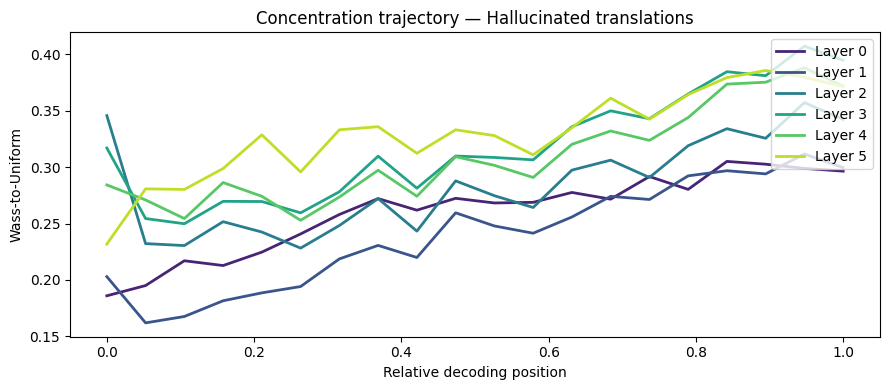}
        \caption{Hallucinated translations.}
    \end{subfigure}
    \caption{Concentration trajectory split by quality group (Fairseq
    DE-EN). Correct translations (left) exhibit the two-phase structure
    with pronounced early dip and layer fan-out. Hallucinated translations
    (right) show compressed, elevated trajectories from step~1 with no
    exploratory phase.}
    \label{fig:concentration-trajectory}
\end{figure}

\medskip
\noindent\textbf{Finding 4.} \textit{Normal decoding follows a two-phase
trajectory: an early exploratory phase (${\sim}$first $25\%$ of the
sequence) followed by a commitment phase of rising concentration and
decaying attention shift. Hallucinated translations lack the exploratory
phase entirely, beginning in a statically concentrated state from step~1
-- consistent with the elevated mean concentration, suppressed step-OT,
and near-zero routing consistency observed in
Section 4.2.}

\medskip
\noindent\textbf{Main takeaway.} OT strongly detects the failure mode
where the decoder detaches from the source. Detection power degrades with
hallucination severity -- from full-unsupport (AUROC $0.957$ with RC)
through strong-unsupport ($0.672$) to repetitions ($0.568$) -- tracking
the degree to which source routing is disrupted.

\section{Experiments II: Transfer to Summarisation}
\label{sec:summ}

\subsection{Faithfulness Detection}

Table~\ref{tab:aggrefact} reports AggreFact results.
The unsupervised OT detector (T5-base) reaches 57.4\% BAcc on average -- above the
50\% random baseline but 14.7 points below supervised MiniCheck-Flan-T5-L.
Flan-T5-large yields a marginally different profile (55.6/61.4\%), with a higher
XSum score consistent with stronger encoders producing more discriminative
concentration signals on abstractive data.

\begin{table*}[ht]
\centering
\caption{Faithfulness detection on AggreFact~\citet{tang2023aggrefact}
         (Balanced Accuracy, \%). Threshold chosen by sweep on the evaluation set.}
\label{tab:aggrefact}
\begin{tabular}{lcccc}
\toprule
Model & CNN & XSum & Avg & Supervision \\
\midrule
OT Detector (T5-base)                         & 57.2 & 57.6 & 57.4 & None \\
OT Detector (Flan-T5-large)                   & 55.6 & 61.4 & 58.5 & None \\
MiniCheck-Flan-T5-L & 69.9 & 74.3 & 72.1 & Supervised (0.8B) \\
Random baseline                               & 50.0 & 50.0 & 50.0 &  --   \\
\bottomrule
\end{tabular}
\end{table*}

A cross-dataset logistic regression (OT features from CNN, tested on XSum) achieves
44.2\% BAcc -- below chance -- confirming that OT features do not generalise across
extractive and abstractive regimes~\citet{maynez2020faithfulness}.

\subsection{Layer-wise Concentration Profile}
\label{sec:conc}

\begin{figure}[ht]
\centering
\includegraphics[width=0.75\linewidth]{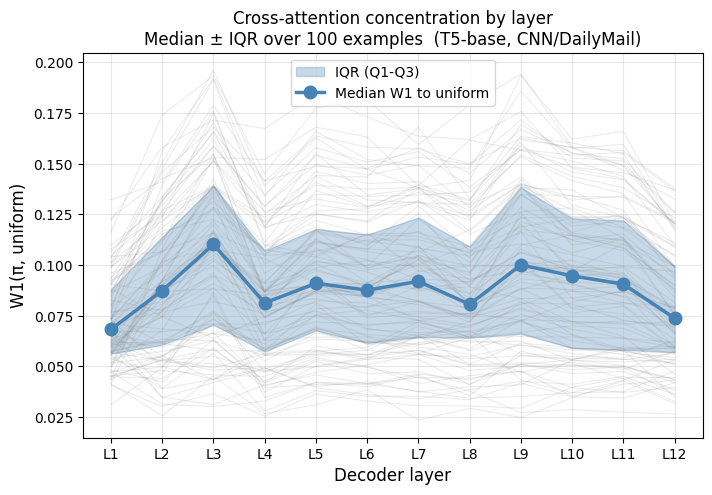}
\caption{Cross-attention concentration $c(\ell)$ by decoder layer
         (T5-base, $N{=}100$). Shaded: IQR. Blue: median. L3 is the global peak.}
\label{fig:conc_profile}
\end{figure}

The cross-attention concentration profile is non-monotonic (Figure~\ref{fig:conc_profile},
Table~\ref{tab:conc}): Layer~3 is the global peak (median $W_1=0.1612$), with
partial relaxation in L4 and secondary elevations at L5, L7, L9.
Layer~12 returns to low concentration (median $W_1=0.0713$).

\begin{table}[ht]
\centering
\caption{Per-layer $W_1(\pi,\mathbf{u})$ for T5-base (median, min, max).
         Global peak marked~$\star$.}
\label{tab:conc}
\begin{tabular}{lccc}
\toprule
Layer & Median $W_1$ & Min & Max \\
\midrule
L1  & 0.0696 & 0.0296 & 0.1065 \\
L2  & 0.0982 & 0.0444 & 0.1509 \\
L3  & \textbf{0.1612}$^{\star}$ & 0.0685 & 0.2404 \\
L4  & 0.1018 & 0.0719 & 0.1462 \\
L5  & 0.1096 & 0.0705 & 0.1445 \\
L6  & 0.0993 & 0.0546 & 0.1532 \\
L7  & 0.0987 & 0.0464 & 0.1554 \\
L8  & 0.1049 & 0.0520 & 0.1535 \\
L9  & 0.1561 & 0.0681 & 0.2021 \\
L10 & 0.1087 & 0.0484 & 0.1776 \\
L11 & 0.0902 & 0.0425 & 0.1711 \\
L12 & 0.0713 & 0.0378 & 0.1755 \\
\bottomrule
\end{tabular}
\end{table}

\begin{figure}[ht]
\centering
\includegraphics[width=\linewidth]{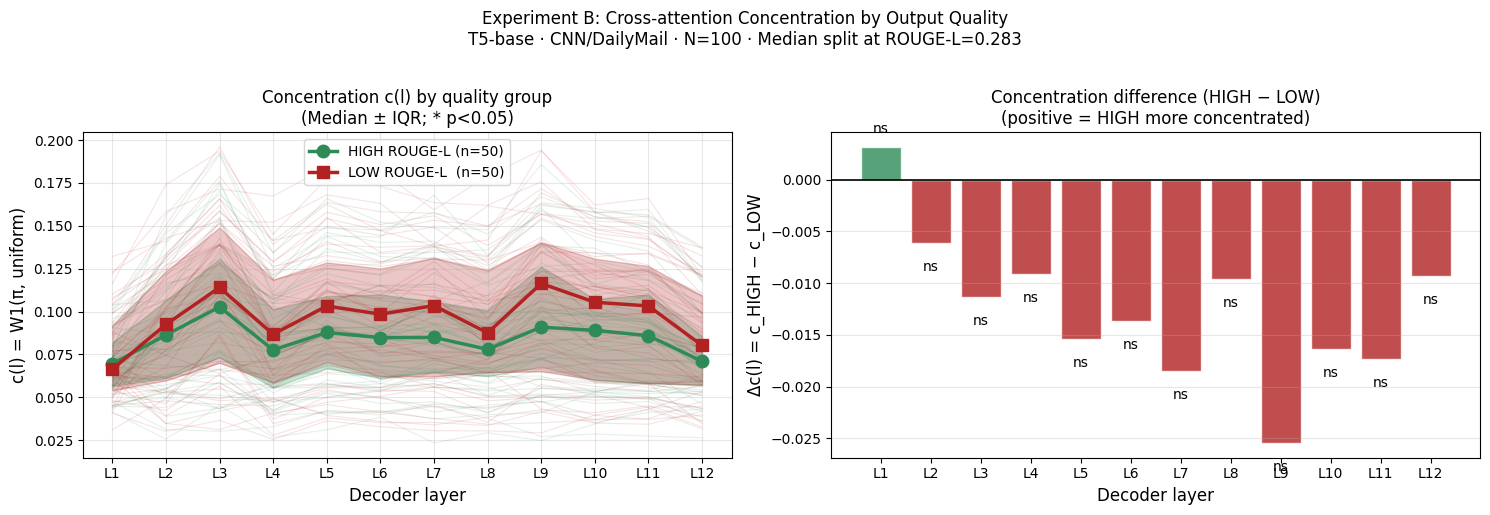}
\caption{Concentration $c(\ell)$ by ROUGE-L quality group (T5-base, $N{=}100$).
         HIGH (green) and LOW (red) are statistically indistinguishable at every
         layer ($p{>}0.60$). Both exhibit the L3 peak.}
\label{fig:conc_highlow}
\end{figure}

Despite this architecture, HIGH and LOW ROUGE-L groups are indistinguishable at
every layer (Figure~\ref{fig:conc_highlow}; Mann-Whitney $U$, $p{>}0.60$ at all
layers): concentration magnitude does not predict summarisation quality.
Pairwise inter-layer distances confirm that L3 is maximally distant from all other
layers (median $D(L1,L3)\approx0.080$), while L9--L11 form a tight cluster
(distances $\approx0.009$--$0.032$), indicating functional redundancy.

\subsection{Layer Ablation}
\label{sec:ablation}

Leave-one-out ablation (Table~\ref{tab:ablation}) on the T5-base baseline
($\text{R-L}=24.94$) converges on the same exceptional layers.
L12 is the single most critical layer ($\Delta\text{R-L}=-0.96$); L5, L7, L11
yield mild improvements when ablated ($+0.59$ to $+0.88$), consistent with the
OT clustering of L9--L11.
Cumulative ablation reveals a critical threshold:
\begin{equation}
  \Delta\mathrm{ROUGE\text{-}L}
  \;=\;
  \begin{cases}
    -0.14 & \text{ablate L1} \\
    -0.07 & \text{ablate L1--L2} \\
    -4.47 & \text{ablate L1--L3}
  \end{cases}
  \label{eq:cumulative}
\end{equation}
The sharp collapse at L1--L3 establishes these layers as a jointly indispensable
early context-encoding block, directly reinforcing the L3 concentration peak.

\begin{table}[ht]
\centering
\caption{Selected leave-one-out ablation results (T5-base, $N{=}50$).
         $\Delta$R-L $=$ ablated $-$ baseline $(24.94)$.}
\label{tab:ablation}
\begin{tabular}{lccc}
\toprule
Layer ablated & ROUGE-L & $\Delta$ R-L & Effect \\
\midrule
L2  & 24.30 & $-0.64$ & Harmful \\
L3  & 24.49 & $-0.45$ & Harmful \\
L5  & 25.53 & $+0.59$ & Beneficial \\
L7  & 25.51 & $+0.57$ & Beneficial \\
L11 & 25.82 & $+0.88$ & Beneficial (max) \\
L12 & 23.98 & $-0.96$ & Critical (max drop) \\
\bottomrule
\end{tabular}
\end{table}

\begin{figure}[ht]
\centering
\includegraphics[width=\linewidth]{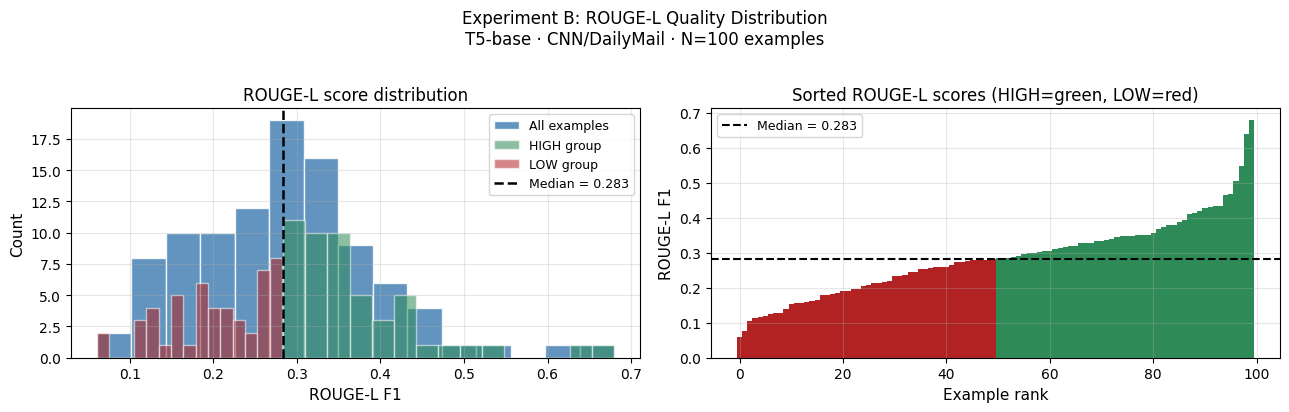}
\caption{ROUGE-L distribution for T5-base ($N{=}100$). Left: histogram (median
         $=0.212$). Right: sorted scores; HIGH\,=\,green, LOW\,=\,red.}
\label{fig:rouge_base}
\end{figure}

\paragraph{Summary.}
The OT detector transfers to summarisation with above-chance performance;
the gap to supervised methods is systematic.
The concentration profile and ablation findings are mutually consistent:
L3 and L12 are structurally and functionally exceptional, while L9--L11 are
redundant.
OT features do not generalise across extractive/abstractive regimes.

% --  --  --  --  --  --  --  --  --  --  --  --  --  --  --  --  --  --  --  --  --  --  --  --  --  --  --  --  -- -
%	5. DISCUSSION
% --  --  --  --  --  --  --  --  --  --  --  --  --  --  --  --  --  --  --  --  --  --  --  --  --  --  --  --  -- -
\section{Discussion}

\subsection{Retrieval Failure vs.\ Content Misuse}

The NMT success of \citet{guerreiro2022} rests on a geometric signal:
hallucinating decoders concentrate attention on irrelevant source tokens
(punctuation, EOS), producing anomalously large $W_1(\pi^{(\ell,t)}, \mathbf{u})$.
In abstractive summarisation, the failure mode is different: an unfaithful summary
may still exhibit concentrated, correctly-targeted attention -- the error occurs
downstream of retrieval, in how the model processes what it attends to.
By construction, the OT score detects retrieval failure only; it is
blind to content misuse.

The dataset-level pattern in Table~\ref{tab:aggrefact} is consistent with this.
CNN/DailyMail summaries are largely extractive, so unfaithful ones may genuinely
exhibit more diffuse attention; XSum is highly abstractive, making the OT signal
noisier even for faithful summaries. The Fairseq results quantify the spectrum
directly: WTU AUROC degrades from $0.937$ (full-unsupport, pure retrieval failure)
to $0.629$ (strong-unsupport, partial source engagement). The AggreFact result
($57.4\%$ BAcc) falls below even the strong-unsupport figure, placing abstractive
faithfulness failures at the far end of this spectrum.

\subsection{OT as an Interpretability Tool}

Despite limited quality prediction, OT metrics provide a principled lens on
Transformer architecture. The L3 concentration peak, its maximum inter-layer
routing divergence, and the functional redundancy of L9--L11 form a coherent
picture of T5-base decoder organisation -- independently confirmed by ablation:
OT-anomalous layers (L3, L12) are ablation-critical; OT-redundant layers (L9--L11)
are ablation-neutral or beneficial. That the structural pivot is L2 in the
six-layer Fairseq model and L3 in twelve-layer T5-base suggests OT localises
functionally exceptional layers consistently across architectures.

A LoRA analysis shows only 0.132\% of T5-base parameters are
adjusted during task adaptation, confirming that the identified attention patterns
are intrinsic to the pre-trained architecture, not task-specific artifacts.

%% ============================================================
\section{Conclusion}

We extended the OT-based detector of \citet{guerreiro2022} to abstractive
summarisation, evaluating on 1,116 AggreFact examples \citet{tang2023aggrefact}.
The unsupervised detector achieves 57.4\% average BAcc -- above chance but
substantially below supervised MiniCheck \citet{tang2023minicheck} (72.1\%).
The gap is principled: OT on cross-attention detects retrieval failure but not
content misuse, and abstractive faithfulness failures are predominantly of the
latter type \citet{maynez2020faithfulness}.

Structural analysis of T5-base \citet{raffel2020} reveals consistent decoder
organisation confirmed across independent methods: Layer~3 is the most selective
and routing-distinct; Layers~L1--L3 form an indispensable early context-encoding
block (Eq.~\ref{eq:cumulative}); Layer~12 is the single most critical for
generation quality. Complementary analysis on the Fairseq DE-EN corpus achieves
AUROC $0.946$ for fully-detached hallucinations (RC: $0.957$), with detection
concentrated in L1--L4 and L5 anti-predictive for subtler types. WTU and WTD are
complementary detectors specialised by hallucination type in consistent,
interpretable ways.

Together, the results support a unified picture: OT on cross-attention is a
reliable detector when the failure mode is source disengagement, a principled
interpretability tool regardless of task, and fundamentally limited when
faithfulness failures occur downstream of attention.

Future work should explore token-level (non-head-averaged) OT signals, larger
instruction-tuned models where attention detachment may be a cleaner failure signal,
and combinations with NLI-based detectors \citet{kryscinski2020evaluating}.

\nocite{github}

\bibliography{report}

@inproceedings{guerreiro2022,
  author    = {Guerreiro, Nuno M. and Martins, Andr{\'e} F. T. and Mariet, Zelda},
  title     = {Optimal Transport for Unsupervised Hallucination Detection in Neural Machine Translation},
  booktitle = {Proceedings of the 61st Annual Meeting of the Association for Computational Linguistics (ACL 2023)},
  year      = {2023},
  url       = {https://arxiv.org/abs/2212.09631}
}

@inproceedings{tang2023aggrefact,
  author    = {Tang, Liyan and Goyal, Tanya and Fabbri, Alexander and Laban, Philippe and Xu, Jiacheng and Koncel-Kedziorski, Rashmi and Choi, Eunsol and Nenkova, Ani and McKeown, Kathleen},
  title     = {Understanding Factual Errors in Summarization: Errors, Summarizers, Datasets, Error Detectors},
  booktitle = {Proceedings of the 61st Annual Meeting of the Association for Computational Linguistics (ACL 2023)},
  year      = {2023},
  url       = {https://arxiv.org/abs/2205.12854}
}

@inproceedings{tang2023minicheck,
  author    = {Tang, Liyan and Laban, Philippe and McKeown, Kathleen},
  title     = {{MiniCheck}: Efficient Fact-Checking of {LLMs} on Grounding Documents},
  booktitle = {Proceedings of the 2024 Conference on Empirical Methods in Natural Language Processing (EMNLP 2024)},
  year      = {2024},
  url       = {https://arxiv.org/abs/2404.10774}
}

@article{raffel2020,
  author    = {Raffel, Colin and Shazeer, Noam and Roberts, Adam and Lee, Katherine and Narang, Sharan and Matena, Michael and Zhou, Yanqi and Li, Wei and Liu, Peter J.},
  title     = {Exploring the Limits of Transfer Learning with a Unified Text-to-Text Transformer},
  journal   = {Journal of Machine Learning Research},
  volume    = {21},
  number    = {140},
  pages     = {1--67},
  year      = {2020},
  url       = {https://arxiv.org/abs/1910.10683}
}

@inproceedings{vaswani2017attention,
  author    = {Vaswani, Ashish and Shazeer, Noam and Parmar, Niki and Uszkoreit, Jakob and Jones, Llion and Gomez, Aidan N. and Kaiser, {\L}ukasz and Polosukhin, Illia},
  title     = {Attention Is All You Need},
  booktitle = {Advances in Neural Information Processing Systems (NeurIPS)},
  volume    = {30},
  year      = {2017},
  url       = {https://arxiv.org/abs/1706.03762}
}

@inproceedings{maynez2020faithfulness,
  author    = {Maynez, Joshua and Narayan, Shashi and Bohnet, Bernd and McDonald, Ryan},
  title     = {On Faithfulness and Factuality in Abstractive Summarization},
  booktitle = {Proceedings of the 58th Annual Meeting of the Association for Computational Linguistics (ACL 2020)},
  year      = {2020},
  url       = {https://arxiv.org/abs/2005.00661}
}

@inproceedings{kryscinski2020evaluating,
  author    = {Kry{\'s}ci{\'n}ski, Wojciech and McCann, Bryan and Xiong, Caiming and Socher, Richard},
  title     = {Evaluating the Factual Consistency of Abstractive Text Summarization},
  booktitle = {Proceedings of the 2020 Conference on Empirical Methods in Natural Language Processing (EMNLP 2020)},
  year      = {2020},
  url       = {https://arxiv.org/abs/1910.12840}
}

@article{peyre2019computational,
  author    = {Peyr{\'e}, Gabriel and Cuturi, Marco},
  title     = {Computational Optimal Transport},
  journal   = {Foundations and Trends in Machine Learning},
  volume    = {11},
  number    = {5--6},
  pages     = {355--607},
  year      = {2019},
  url       = {https://arxiv.org/abs/1803.00567}
}

@misc{github,
  title  = {Code implementation and analytics},
  author = {Anonymous},
  year   = {2026},
  url    = {https://anonymous.4open.science/r/Layer_Resolved_Optimal_Transport}
}
\bibliographystyle{icml2026}

\end{document}